\documentclass{article}
 \usepackage{ijcai15}
 \usepackage{times}
 \usepackage{helvet}
 \usepackage{courier}
\usepackage{amsmath}
\newtheorem{THEOREM}{Theorem}[section]
\newenvironment{theorem}{\begin{THEOREM} \hspace{-.85em} {\bf :} }%
                        {\end{THEOREM}}
\newtheorem{LEMMA}[THEOREM]{Lemma}
\newenvironment{lemma}{\begin{LEMMA} \hspace{-.85em} {\bf :} }%
                      {\end{LEMMA}}
\newtheorem{COROLLARY}[THEOREM]{Corollary}
\newenvironment{corollary}{\begin{COROLLARY} \hspace{-.85em} {\bf :} }%
                          {\end{COROLLARY}}
\newtheorem{PROPOSITION}[THEOREM]{Proposition}
\newenvironment{proposition}{\begin{PROPOSITION} \hspace{-.85em} {\bf :} }%
                            {\end{PROPOSITION}}
\newtheorem{DEFINITION}[THEOREM]{Definition}
\newenvironment{definition}{\begin{DEFINITION} \hspace{-.85em} {\bf :} \rm}%
                            {\end{DEFINITION}}
\newtheorem{CLAIM}[THEOREM]{Claim}
\newenvironment{claim}{\begin{CLAIM} \hspace{-.85em} {\bf :} \rm}%
                            {\end{CLAIM}}
\newtheorem{EXAMPLE}[THEOREM]{Example}
\newenvironment{example}{\begin{EXAMPLE} \hspace{-.85em} {\bf :} \rm}%
                            {\end{EXAMPLE}}
\newtheorem{REMARK}[THEOREM]{Remark}
\newenvironment{remark}{\begin{REMARK} \hspace{-.85em} {\bf :} \rm}%
                            {\end{REMARK}}
\newcommand{\thm}{\begin{theorem}}
\newcommand{\lem}{\begin{lemma}}
\newcommand{\pro}{\begin{proposition}}
\newcommand{\dfn}{\begin{definition}}
\newcommand{\rem}{\begin{remark}}
\newcommand{\xam}{\begin{example}}
\newcommand{\cor}{\begin{corollary}}
\newcommand{\prf}{\noindent{\bf Proof:} }
\newcommand{\ethm}{\end{theorem}}
\newcommand{\elem}{\end{lemma}}
\newcommand{\epro}{\end{proposition}}
\newcommand{\edfn}{\bbox\end{definition}}
\newcommand{\erem}{\bbox\end{remark}}
\newcommand{\exam}{\bbox\end{example}}
\newcommand{\ecor}{\end{corollary}}
\newcommand{\eprf}{\bbox\vspace{0.1in}}
\newcommand{\beqn}{\begin{equation}}
\newcommand{\eeqn}{\end{equation}}
\newcommand{\bbox}{\vrule height7pt width4pt depth1pt}

\newcommand{\clm}{\begin{claim}}
\newcommand{\eclm}{\end{claim}}
\newcommand{\sat}{\models}

\newcommand{\union}{\cup}
\newcommand{\inter}{\cap}
\newcommand{\FF}{{\bf F}}

\renewcommand{\phi}{\varphi}
\newcommand{\F}{{\cal F}}

\newcommand{\R}{{\cal R}}

\newcommand{\U}{{\cal U}}
\newcommand{\V}{{\cal V}}

\renewcommand{\>}{\rangle}

\newcommand{\ol}{\setlength{\itemsep}{0pt}\begin{enumerate}}
\newcommand{\eol}{\end{enumerate}\setlength{\itemsep}{-\parsep}}
\newcommand{\ul}{\setlength{\itemsep}{0pt}\begin{itemize}}
\newcommand{\dl}{\setlength{\itemsep}{0pt}\begin{description}}
\newcommand{\edl}{\end{description}\setlength{\itemsep}{-\parsep}}
\newcommand{\eul}{\end{itemize}\setlength{\itemsep}{-\parsep}}

\newcommand{\BS}{B^{\scriptscriptstyle \cS}}

\newcommand{\commentout}[1]{}

\newcommand{\bi}{\begin{itemize}}
\newcommand{\ei}{\end{itemize}}
\newcommand{\be}{\begin{enumerate}}
\newcommand{\ee}{\end{enumerate}}

\renewcommand{\S}{{\cal S}}
\newcommand{\NP}{\mbox{{\it NP}}}
\newcommand{\DP}{D^P}
\newcommand{\SH}{\mbox{{\it SH}}}
\newcommand{\BH}{\mbox{{\it BH}}}
\newcommand{\BT}{\mbox{{\it BT}}}
\renewcommand{\BS}{\mbox{{\it BS}}}
\newcommand{\ST}{\mbox{{\it ST}}}
\newcommand{\RT}{\mathit{RT}}
\renewcommand{\FF}{\mathit{FF}}
\newcommand{\ML}{\mathit{MD}}
\newcommand{\Lan}{L}
\newcommand{\zug}[1]{\langle #1  \rangle}
\newcommand{\shortv}{\commentout}
\newcommand{\fullv}[1]{#1}
\newcommand{\journal}[1]{}
\newcommand{\Lsingle}{L_{\mbox{\small AC2}}}
\newcommand{\Lmin}{L_{\mbox{\small AC3}}}
\newcommand{\pair}[1]{\langle #1  \rangle}
\renewcommand{\citeyear}{\shortcite}
\newcommand{\LB}{\mbox{{\it LB}}}
\newcommand{\RB}{\mbox{{\it RB}}}

\begin{document}

\title{A Modification of the Halpern-Pearl Definition of Causality}
\author{Joseph Y. Halpern%
\thanks{Work supported in part by NSF grants 
IIS-0911036, and CCF-1214844, by AFOSR grants
FA9550-09-1-0266 and FA9550-12-1-0040, and by ARO
grant W911NF-14-1-0017.}\\
   Cornell University\\
   Computer Science Department\\
   Ithaca, NY 14853\\
   halpern@cs.cornell.edu\\
   http://www.cs.cornell.edu/home/halpern}
\maketitle

\begin{abstract}
The original Halpern-Pearl definition of causality \cite{HPearl01a}
was updated in the journal version of the paper \cite{HP01b}
to deal with some problems pointed out by Hopkins and Pearl
\citeyear{HopkinsP02}.  Here the definition is modified yet again, in a way
that (a) leads to a simpler definition, (b) handles the problems
pointed out by Hopkins and Pearl, and many others, (c) gives
reasonable answers (that agree with those of the original and updated
definition) in the 
standard problematic examples of causality, and (d) has lower complexity 
than either the original or updated definitions.
%The
%implications of the modification for the notions of responsibility and
%blame \cite{ChocklerH03} is also considered.
\end{abstract}

\section{Introduction}
Causality plays a central role in the way people structure the world.
People constantly seek causal explanations for their observations.
Philosophers have typically distinguished two notions of causality,
which they have called \emph{type causality} (sometimes called
\emph{general causality}) and \emph{actual causality} (sometimes
called  \emph{token causality} or \emph{specific causality}). Type
causality is perhaps 
what
 scientists are most concerned with.  These are general
statements, such as ``smoking causes lung cancer'' and ``printing money
causes inflation''. 
By way of contrast, actual causality focuses on particular events:
``the fact that David smoked like a chimney for 30 years caused him to
get cancer last year'';  
``the car's faulty brakes caused the accident (not the
pouring rain or the driver's drunkenness)''.   Here I focus on actual causality.

Despite the fact that the use of causality is ubiquitous, and that it plays
a key role in science and in the determination of legal cases (among
many other things), finding a good definition of actual causality has proved
notoriously difficult.  Most recent definitions of actual causality, going
back to the work of Lewis \citeyear{lewis73a}, involve
counterfactuals.  The idea is that $A$ is a cause of $B$ if, had $A$
not happened, $B$ would not have happened.  This is the standard
``but-for'' test used in the law: but for $A$, $B$ would not have occurred.

However, as is well known, the but-for test is not always sufficient
to determine causality.  Consider the following well-known example,
taken from \cite{HallP03}:
\begin{quote}
Suzy and Billy both pick up rocks
and throw them at  a bottle.
Suzy's rock gets there first, shattering the
bottle.  Since both throws are perfectly accurate, Billy's would have
shattered the bottle
had it not been preempted by Suzy's throw.
\end{quote}
Here the but-for test fails.  Even if Suzy hadn't thrown, the bottle
would have shattered.  Nevertheless, we want to call Suzy's throw a
cause of the bottle shattering.
%, and not Billy's throw.

 Halpern and Pearl \citeyear{HPearl01a} introduced
a definition using \emph{structural equations} that has proved quite
influential.  In the structural-equations approach, the world is
assumed to be characterized by the values of a collection of
variables.  In this example, we can use binary variable $\ST$ for
``Suzy throws'' 
($\ST = 1$ if Suzy throws; $\ST=0$ if she doesn't),  $\BT$ for ``Billy
throws'', and $\BS$ for ``bottle shatters''. To show that $\ST=1$ is a
cause of $\BS=1$, the Halpern-Pearl (henceforth HP) 
%joe1
definition 
allows us to
consider a situation where Billy does not throw (i.e., $\BT$ is set to
0).  Under that contingency, the but-for definition works just right:
if Suzy doesn't throw, the bottle doesn't shatter, and if Suzy throws,
the bottle does shatter.  

There is an obvious problem with this approach: it can also be used to
show that Billy's throw is a cause of the bottle shattering, which we
do not want.  Halpern and Pearl deal with this problem by adding extra
variables to the story; this is needed to make it clear that Suzy and
Billy play asymmetric roles.  Specifically, they add variables $\SH$
(for ``Suzy hits the bottle'') and $\BH$ (for ``Billy hits the bottle'');
in the actual situation, $\SH=1$ and $\BH=0$.  By putting appropriate 
restrictions on which contingencies can be considered, they show that
the HP definition does indeed allow us to conclude that $\ST = 1$ is a
cause of $\BS=1$, and $\BT=1$ is not.  (See Section~\ref{sec:examples} for details.)

However, the question of which contingencies can be considered turns
out to be subtle.  Hopkins and Pearl \citeyear{HopkinsP02} gave an
example where the original HP definition gave arguably inappropriate
results; it was updated in the journal version of the paper
\cite{HP01b} in a way that deals with this example.
%\fullv{
%\footnote{Interestingly, the Hopkins and Pearl example can be dealt
% with by adding extra variables, just as the Suzy-Billy example
% \cite{Hal44}.  However, there are other examples where it seems that
% the modified definition gives more reasonable results than the
% original definition \cite{BBCOT12}.}
%}
Further counterexamples were given to the updated definition (see,
for example, \cite{Hall07,Hiddleston05,Weslake11}).  By and large,
these examples can be dealt with by taking into account considerations
of normality and defaults \cite{Hal39,HH11} or by adding extra variables to
the model (see \cite{Hal44}).  But these approaches do not always seem
so satisfactory.

In this paper, I further modify the HP definition, by placing more
stringent restrictions on the contingencies that can be considered.
Roughly speaking, when we consider various contingencies, I do not
allow the values of variables other than that of the putative cause(s) 
to be changed; I simply allow values to be
frozen at their actual values.  Thus, for example, in the Suzy-Billy
example, I do not consider the contingency where Billy does not throw
(since that would involve change the value of $\BT$ from its actual
value).  But I do allow $\BH$ to be frozen at its actual value of 0
when considering the possibility that Suzy does not throw.
This results in a definition that is significantly simpler than the HP
definition, deals well with all the standard examples in the
literature, and deals with some of the problem cases better than the
HP definition.  In addition, the complexity of computing causality is
$\Delta^p$, simpler than that of either the original HP definition 
or the modification proposed by HP (cf.~\cite{ACHI14,EL01}.  

The rest of this paper is organized as follows.  In the next section,
I review the original and updated HP definitions, and introduce the
modification.  In Section~\ref{sec:examples}, I compare the
definitions in various examples, and show that the modified 
definition gives more reasonable results than the
original and updated definitions.  
\fullv{In Section~\ref{sec:compare}, I compare the modified definition
with definitions given by Hitchcock~\citeyear{hitchcock:99}, Hall
\citeyear{Hall07}, and Pearl~\citeyear{pearl:2k}.}
In Section~\ref{sec:complexity}, I
consider the complexity of computing causality under the modified
definition.  I conclude in Section~\ref{sec:conclusion}.
\shortv{Proofs of theorems and further examples can be found in the
  full paper, available at
%anonymously at   https://drive.google.com/file/d/
%0B\_kL2mcLOlvxMmRCTjNXWFY4QWc/view?usp=sharing.} 
% 0B_kL2mcLOlvxMmRCTjNXWFY4QWc&authuser=0
%https://drive.google.com/open?id=0B_kL2mcLOlvxMmRCTjNXWFY4QWc&authuser=0
www.cs.cornell.edu/home/halpern/modified-HPdef.pdf.}

\section{The HP definition(s) and the modified
  definition}\label{sec:definitions} 

In this section, I review the HP definition of causality and introduce
the modified definition.
The reader is encouraged to consult
\cite{HP01b} for further details and intuition regarding the HP definition.
The exposition of the review material is largely taken from \cite{Hal39}.

\subsection{Causal structures}
The HP approach assumes that the world is described in terms of 
%joe*
%random
variables and their values.  
Some variables may have a causal influence on others. This
influence is modeled by a set of {\em structural equations}.
It is conceptually useful to split the variables into two
sets: the {\em exogenous\/} variables, whose values are
determined by 
factors outside the model, and the
{\em endogenous\/} variables, whose values are ultimately determined by
the exogenous variables.  
For example, in a voting scenario, we could have endogenous variables
that describe what the voters actually do (i.e., which candidate they
vote for), exogenous variables 
that describe the factors
that determine how the voters vote, and a
variable describing the outcome (who wins).  The structural equations
describe how the outcome is determined (majority rules; a candidate
wins if $A$ and at least two of $B$, $C$, $D$, and $E$ vote for him;
etc.).

Formally, a \emph{causal model} $M$
is a pair $(\S,\F)$, where $\S$ is a \emph{signature}, which explicitly
lists the endogenous and exogenous variables  and characterizes
their possible values, and $\F$ defines a set of \emph{modifiable
structural equations}, relating the values of the variables.  
A signature $\S$ is a tuple $(\U,\V,\R)$, where $\U$ is a set of
exogenous variables, $\V$ is a set 
of endogenous variables, and $\R$ associates with every variable $Y \in 
\U \union \V$ a nonempty set $\R(Y)$ of possible values for 
$Y$ (that is, the set of values over which $Y$ {\em ranges}).  
For simplicity, I assume here that $\V$ is finite, as is $\R(Y)$ for
every endogenous variable $Y \in \V$.
$\F$ associates with each endogenous variable $X \in \V$ a
function denoted $F_X$ such that $F_X: (\times_{U \in \U} \R(U))
\times (\times_{Y \in \V - \{X\}} \R(Y)) \rightarrow \R(X)$.
This mathematical notation just makes precise the fact that 
$F_X$ determines the value of $X$,
given the values of all the other variables in $\U \union \V$.
If there is one exogenous variable $U$ and three endogenous
variables, $X$, $Y$, and $Z$, then $F_X$ defines the values of $X$ in
terms of the values of $Y$, $Z$, and $U$.  For example, we might have 
$F_X(u,y,z) = u+y$, which is usually written as
$X = U+Y$.   Thus, if $Y = 3$ and $U = 2$, then
$X=5$, regardless of how $Z$ is set.%
\footnote{The fact that $X$ is assigned  $U+Y$ (i.e., the value
of $X$ is the sum of the values of $U$ and $Y$) does not imply
that $Y$ is assigned $X-U$; that is, $F_Y(U,X,Z) = X-U$ does not
necessarily hold.}    

The structural equations define what happens in the presence of external
interventions. 
Setting the value of some variable $X$ to $x$ in a causal
model $M = (\S,\F)$ results in a new causal model, denoted $M_{X
\gets x}$, which is identical to $M$, except that the
equation for $X$ in $\F$ is replaced by $X = x$.

Following \cite{HP01b}, I restrict attention here to what are called {\em
recursive\/} (or {\em acyclic\/}) models.  This is the special case
where there is some total ordering $\prec$ of the endogenous variables
(the ones in $\V$) 
such that if $X \prec Y$, then $X$ is independent of $Y$, 
that is, $F_X(\ldots, y, \ldots) = F_X(\ldots, y', \ldots)$ for all $y, y' \in
\R(Y)$.  Intuitively, if a theory is recursive, there is no
feedback.  If $X \prec Y$, then the value of $X$ may affect the value of
$Y$, but the value of $Y$ cannot affect the value of $X$.
It should be clear that if $M$ is an acyclic  causal model,
then given a \emph{context}, that is, a setting $\vec{u}$ for the
exogenous variables in $\U$, there is a unique solution for all the
equations.  We simply solve for the variables in the order given by
$\prec$. The value of the variables that come first in the order, that
is, the variables $X$ such that there is no variable $Y$ such that $
Y\prec X$, depend only on the exogenous variables, so their value is
immediately determined by the values of the exogenous variables.  
The values of variables later in the order can be determined once we have
determined the values of all the variables earlier in the order.

%It is sometimes helpful to represent a causal model graphically. 
%Each node in the graph corresponds to one variable in the model.
%An arrow from one node to another indicates that
%the former variable figures as a nontrivial argument in the equation 
%for the latter.   The graphical representation is useful for visualizing
%causal models, and will be used in the next section.

\subsection{A language for reasoning about causality}\label{sec:lang}
To define causality carefully, it is useful to have a language to reason
about causality.
Given a signature $\S = (\U,\V,\R)$, a \emph{primitive event} is a
formula of the form $X = x$, for  $X \in \V$ and $x \in \R(X)$.  
A {\em causal formula (over $\S$)\/} is one of the form
$[Y_1 \gets y_1, \ldots, Y_k \gets y_k] \phi$,
where
\fullv{
\begin{itemize}
\item}
$\phi$ is a Boolean
combination of primitive events,
\fullv{\item} $Y_1, \ldots, Y_k$ are distinct variables in $\V$, and
\fullv{\item} $y_i \in \R(Y_i)$.
\fullv{\end{itemize}}
Such a formula is
abbreviated
as $[\vec{Y} \gets \vec{y}]\phi$.
The special
case where $k=0$
is abbreviated as
$\phi$.
Intuitively,
$[Y_1 \gets y_1, \ldots, Y_k \gets y_k] \phi$ says that
%$\phi(\vec{u})$ holds in the counterfactual world that would arise if
$\phi$ would hold if
$Y_i$ were set to $y_i$, for $i = 1,\ldots,k$.

A causal formula $\psi$ is true or false in a causal model, given a
context.
As usual, I write $(M,\vec{u}) \sat \psi$  if
the causal formula $\psi$ is true in
causal model $M$ given context $\vec{u}$.
The $\sat$ relation is defined inductively.
$(M,\vec{u}) \sat X = x$ if
the variable $X$ has value $x$
in the
unique (since we are dealing with acyclic models) solution
to
the equations in
$M$ in context $\vec{u}$
(that is, the
unique vector
of values for the exogenous variables that simultaneously satisfies all
equations 
in $M$ 
with the variables in $\U$ set to $\vec{u}$).
The truth of conjunctions and negations is defined in the standard way.
Finally, 
$(M,\vec{u}) \sat [\vec{Y} \gets \vec{y}]\phi$ if 
$(M_{\vec{Y} = \vec{y}},\vec{u}) \sat \phi$.
%I write $M \sat \phi$ if $(M,\vec{u}) \sat \phi$ for all contexts $\vec{u}$.

\subsection{The definition of causality}\label{sec:causalitydef}
The original HP definition, the updated HP definition, and the
modification I introduce here all have three clauses, denoted AC1,
AC2, and AC3.  The definitions differ only in AC2.  
AC1 and AC3 are  simple and straightforward; all the 
``heavy lifting'' is done by AC2.  In all cases, the definition of
causality, like the definition of truth discussed in
Section~\ref{sec:lang}, is relative to a model and a context.

\dfn\label{actcaus}
$\vec{X} = \vec{x}$ is an \emph{actual cause of $\phi$ in
$(M, \vec{u})$} if the following
three conditions hold:
\begin{description}
\item[{\rm AC1.}]\label{ac1} $(M,\vec{u}) \sat (\vec{X} = \vec{x})$ and 
$(M,\vec{u}) \sat \phi$.
%(That is, for $\vec{X} = \vec{x}$ to be a cause of $\phi$, both $\vec{X}
\item[{\rm AC2.}]\label{ac2} Discussed below.
\item[{\rm AC3.}] \label{ac3}
$\vec{X}$ is minimal; no subset of $\vec{X}$ satisfies
conditions AC1 and AC2.
\label{def3.1}  %%changed from {def2.2}
\end{description}
\end{definition}

AC1 just says that $\vec{X}=\vec{x}$ cannot
be considered a cause of $\phi$ unless both $\vec{X} = \vec{x}$ and
$\phi$ actually happen.  AC3 is a minimality condition, which ensures
that only those elements of 
the conjunction $\vec{X}=\vec{x}$ that are essential are
considered part of a cause; inessential elements are pruned.
Without AC3, if dropping a lit cigarette is a cause of a fire
then so is dropping the cigarette and sneezing.

AC2 is the core of the definition.  I start by presenting the original
definition of AC2, taken from \cite{HPearl01a}.   In this definition, 
AC2 consists of
two parts, AC2(a) and AC2(b).  
AC2(a) is a necessity condition.  It says that for $X=x$
to be a cause of $\phi$, there must be a setting $x'$ such that if $X$
is set to $x'$, $\phi$ would not have occurred.  This is the but-for
clause; but for the fact that $X=x$ occurred, $\phi$ would not have
occurred.  As we saw in the Billy-Suzy rock-throwing example,
the naive but-for clause will not
suffice.  The original HP definition allows us to apply the
but-for definition to contingencies 
where some variables are set to values other than those that they take
in the actual situation.  For example, in the case of Suzy and Billy,
we consider a contingency where Billy does not throw.  

\begin{description}
\item[{\rm AC2(a).}] There is a partition of $\V$ (the set of endogenous variables) into two
disjoint subsets $\vec{Z}$ and $\vec{W}$  (so that $\vec{Z} \inter
\vec{W} = \emptyset$) 
with $\vec{X} \subseteq \vec{Z}$ and a
setting $\vec{x}'$ and $\vec{w}$ of the variables in $\vec{X}$ and
$\vec{W}$, respectively, such that
$$(M,\vec{u}) \sat [\vec{X} \gets \vec{x}',
\vec{W} \gets \vec{w}]\neg \phi.$$
\end{description}
%We can think of the variables in $\vec{Z}$ as making up the ``causal
%path'' from $\vec{X}$ to $\phi$.  Intuitively, changing the value of
%some variable in $X$ results in changing the value(s) of some
%variable(s) in $\vec{Z}$, which results in the values of some
%other variable(s) in $\vec{Z}$ being changed, which finally results in
%the value of $\phi$ changing.  The remaining endogenous variables, the
%ones in $\vec{W}$, are off to the side, so to speak, but may still have
%an indirect effect on what happens.
So AC2(a) says that the but-for condition holds under the contingency 
$\vec{W} = \vec{w}$.

Unfortunately, AC1, AC2(a), and AC3 do not suffice for a good
definition of causality.  In the rock-throwing example, with just AC1,
AC2(a), and AC3, Billy would be a cause of the bottle shattering.  We
need a sufficiency condition to block Billy.  
Roughly speaking, the sufficiency condition
requires that if $\vec{X}$ 
is set to $\vec{x}$, then $\phi$ holds even if $\vec{W}$ is set to
$\vec{w}$ and all the variables in an arbitrary subset 
$\vec{Z}'$ of $\vec{Z}$ are set to their values in the actual context
(where  the
value of a variable $Y$ in the actual context is the value $y$ such that
$(M,u) \sat Y=y$).  Formally, using the notation of AC2(a), we have
\begin{description}
\item[{\rm AC2(b)}.] If $\vec{z}$ is such that 
$(M,\vec{u}) \sat \vec{Z} = \vec{z}$, then, for all subsets $\vec{Z}'$
of $\vec{Z}$, we have
$$ 
(M,\vec{u}) \sat [\vec{X} \gets \vec{x}, \vec{W} \gets \vec{w}, \vec{Z}'
    \gets \vec{z}]\phi.%
\footnote{There is a slight abuse of notation here.  Suppose that 
$\vec{Z} = (Z_1,Z_2)$, $\vec{z} = (1,0)$, and $\vec{Z}' = (Z_1)$.
Then $\vec{Z}' \gets \vec{z}$ is intended to be an abbreviation for
$Z_1 \gets 1$; that is, I am ignoring the second component of
$\vec{z}$ here.
More generally, when I write $\vec{Z}' \gets \vec{z}$,
I am picking out the values in $\vec{z}$ that correspond to the
variables in $\vec{Z}'$, and ignoring those that correspond to the
variables in $\vec{Z} - \vec{Z}'$.  I similarly write $\vec{W}' \gets
\vec{w}$ if $\vec{W}'$ is a subset of $\vec{W}$.    Also note that
although I use the vector notation $\vec{Z}$, I sometimes view
$\vec{Z}$ as a set of variables.}
$$
\end{description}

The updated HP definition \cite{HP01b} strengthens AC2(b) further.
Sufficiency is required to hold if the variables in \emph{any subset}
$\vec{W}'$ of $\vec{W}$ are set to the values
in $\vec{w}$ (in addition to allowing the variables in 
any  subset $\vec{Z}'$ of $\vec{Z}$  
to be set to their values in the actual context).  Formally, 
the following condition AC2(b$^u$) must hold (the ``u''
stands for ``updated''):
\begin{description}
\item[{\rm AC2(b$^u$)}.] If $\vec{z}$ is such that 
$(M,\vec{u}) \sat \vec{Z} = \vec{z}$, then, for all subsets $\vec{W}'$
  of $\vec{W}$ and $\vec{Z}'$ of $\vec{Z}$, we have 
$$(M,\vec{u}) \sat [\vec{X} \gets \vec{x}, 
\vec{W}' \gets \vec{w}, \vec{Z}' \gets \vec{z}]\phi.$$
\end{description}
Requiring sufficiency to hold for all subsets $\vec{W}'$ of
$\vec{W}$ is analogous to requirement in AC2(b) that it hold for all
subsets $\vec{Z}'$ of $\vec{Z}$.  
Some motivation for these requirements is given in the examples in
Section~\ref{sec:examples}.  

The modified definition is motivated by the observation that when we want
to argue that Suzy is the cause of the bottle shattering, and not
Billy, we point out that what actually happened is that Suzy's throw
hit the bottle, while Billy's rock didn't.  That is, what matters is
what happened in the actual situation.  Thus, the only settings of
variables allowed are ones that occurred in the actual situation.
Specifically, the modified definition simplifies AC2(a) by requiring that the
only setting $\vec{w}$ of the variables in $\vec{W}$ that can be considered
is the value of these variables in the actual context.
Here is the modified AC2(a),
which I denote AC2(a$^m$) (the $m$ stands for ``modified''):

\begin{description}
\item[{\rm AC2(a$^m$)}.]  There is a set $\vec{W}$ of variables in $\V$
and a setting $\vec{x}'$ of the variables in $\vec{X}$ such that 
if  $(M,\vec{u}) \sat \vec{W} = \vec{w}$, then
$$(M,\vec{u}) \sat [\vec{X} \gets \vec{x}',
\vec{W} \gets \vec{w}]\neg \phi.$$
\end{description}
Because $\vec{w}$ is the value of the variables in $\vec{W}$ 
in the actual context, AC2(b$^u$) 
follows immediately from AC1 and AC2(a$^m$); so does AC2(b).  
Thus, there is no need for an analogue to AC2(b) in the modified
definition.  Moreover, 
%joe1
%the modified definition also does not need to mention $\vec{Z}$ (although
the modified definition does not need to mention $\vec{Z}$ (although
$\vec{Z}$ can be taken to be the complement of $\vec{W}$).  

For future reference, the tuple $(\vec{W}, \vec{w}, \vec{x}')$ in AC2
is said to be a \emph{witness} to the fact that $\vec{X} = \vec{x}$ is
a cause of $\phi$.  \fullv{(I take the witness to be $(\emptyset, \emptyset,
\vec{x}')$ in the special case that $\vec{W} = \emptyset$.)}
Each conjunct in $\vec{X} = \vec{x}$ is called
\emph{part} of a cause of $\phi$ in context $(M,\vec{u})$.
As we shall see, 
what we think of as causes in natural language often correspond to parts of
causes with the modified HP definition.

%\fullv{
%This idea of fixing variables at their actual values already appears
%in a definition of causality due to Hitchcock \citeyear{hitchcock:99}. 
%However,
%Hitchcock's definition involves a path of variables from the cause to
%the effect, which leads to some nontrivial technical differences
%between my definition and his (see Example~\ref{xam:hitchcock}).
%}

The differences between these definitions will become clearer when I
consider a number of examples in the next section.
For ease of reference, I call the definition satisfying AC2(a) and
AC2(b) the \emph{original HP definition}, the definition satisfying
AC2(a) and AC2(b$^u$) the \emph{updated HP definition}, and the
definition satisfying AC2(a$^m$) the \emph{modified definition}.  
Note that just as there are three versions of AC2, technically, there are 
three corresponding versions of AC3.  For example, in the case of the
modified definition, AC3 should really say ``there is no subset of
$\vec{X}$ satisfying AC1 and AC2(a$^m$)''.  I will not bother writing
out these versions of AC3; I hope that the intent is clear whenever I
refer to AC3.

At this point, ideally, I would prove a theorem showing that some
variant of the HP definition of actual causality is 
is the ``right'' definition of actual causality.  But I know of no way
to argue convincingly that a definition  is the ``right'' one; the
best we can hope to do is to show that it is useful.   As a first
step, I show that all definitions agree in the simplest, and arguably
most common case: but-for causes.
Formally, say that $X=x$ is a \emph{but-for cause of $\phi$ in
  $(M,\vec{u})$} if 
AC1 holds (so that $(M,\vec{u}) \sat X=x \land \phi$) and 
there exists some $x'$ such that $(M,\vec{u}) \sat [X \gets x']\neg\phi$.
Note here I am assuming that the cause is a single conjunct.  

\pro\label{pro:butfor} If $X=x$ is a but-for cause of $Y=y$ in 
  $(M,\vec{u})$, then $X=x$ is a cause of $Y=y$ according to 
all three variants of the HP definition.
\epro
\fullv{
\prf Suppose that $X=x$ is a but-for cause of $Y=y$ and $x'$ is such
that $(M,\vec{u}) \sat [X \gets x']\neg\phi$.  
Then $(\emptyset, \emptyset, x')$ is a witness for $X=x'$ being a
cause of $\phi$ for
  all three variants of the definition.  
Thus, AC2(a) and AC2(a$^m$) hold if we take $\vec{W} = \emptyset$.
Since $(M,\vec{u}) \sat X=x$, if $(M,\vec{u}) \sat \vec{Z} =
\vec{z}$, where $\vec{Z} = \V - \{X\}$, then it is easy to see that 
$(M,\vec{u}) \sat [X \gets x](\vec{Z} - \vec{z})$: setting $X$ to its actual
value does not affect the actual value of any other variable,
since $M_{X \gets x} = M$.  
Similarly, $M_{X \gets X, \vec{Z} \gets \vec{z}} = M$, so 
$(M,\vec{u}) \sat [\vec{X} \gets \vec{x}, \vec{Z}'
    \gets \vec{z}]\phi$ for all subsets $\vec{Z}'$ of $\V - \{X\}$.  
Thus, AC2(b$^o$) holds.  Because $\vec{W} = \emptyset$, AC2(b$^u$)
follows immediately from AC2(b$^o$).  
%Thus, $(\emptyset, \emptyset, x')$ is the witness for
%  all three variants of the definition.  
\eprf
}

Of course, the definitions do not always agree.
\fullv{As the following theorem shows, the}
\shortv{The} modified definition is more
stringent than the original or updated definitions; if $X=x$ is part of
a cause of $\phi$ according to the modified definition,
then it is also a cause according to both the 
original and updated definitions.

\thm\label{thm:connections}
If $X=x$ is part of a cause of $\phi$ in $(M,\vec{u})$
  according to the modified HP definition, then 
$X=x$ is a cause of $\phi$ in $(M,\vec{u})$
  according to both the original and updated HP definitions. 
\ethm

\fullv{\prf See the appendix. \eprf}

\section{Examples}\label{sec:examples}
In this section, I consider how the definitions play out in a number of
examples.  The first example is taken from \cite{HPearl01a}, with
minor variations.

\xam\label{xam:ff} An arsonist drops a lit match in a dry forest and lightning
strikes a tree, setting it on fire.  Eventually the forest burns down.  
We are interested in the cause of the fire.  We can describe the
world using three endogenous variables:
\begin{itemize}
\item $\FF$ for forest fire, where $\FF=1$ if there is a forest fire and
$\FF=0$ otherwise; 
\item $L$ for lightning, where $L=1$ if lightning occurred and $L=0$ otherwise;
\item $\ML$ for match dropped (by arsonist), where $\ML=1$ if the arsonist
dropped a lit match, and $\ML = 0$ otherwise.
\end{itemize}
We also have an exogenous variable $U$ that determines whether the
arsonist drops the match and whether there is lightning.  Take
$\R(U) = \{(i,j): i, j \in \{0,1\}  \}$, where the arsonist drops the
match if $i=1$ and the lightning strikes if $j=1$.  We are interested
in the context $(1,1)$.

Consider two scenarios.  In the first, called the \emph{disjunctive
  scenario}, either the match or the lightning suffice to cause the
fire. In the second, called the \emph{conjunctive scenario}, both are
needed for the forest to burn down.  The scenarios differ in the
equations for $\FF$.  In the model $M_C$ for the conjunctive scenario, we
have the equation $\FF = \min(L,\ML)$ (or $\FF = L \land \ML$, if we
identify binary variables with primitive propositions, where 1 denotes
``true''); in the model $M_D$ for the disjunctive scenario, we have
the equation $\FF = \max(L,\ML)$ (or $\FF = L \lor \ML$).    

In the conjunctive scenario, all the definitions agree that both the
lightning and the arsonist are causes, since each of $L=1$ and $\ML=1$
is a but-for cause of $\FF = 1$ in $(M_C,(1,1))$.
This example also shows that all three
definitions allow for more than one cause of an effect.

In the disjunctive scenario, the original and updated HP definitions
again would call each of $L = 1$ and $\ML=1$ causes. 
I give the argument here for $L=1$.  
Again, the fact that AC1 and AC3 hold is immediate.  For AC2, let
$\vec{Z} = \{L,\FF\}$ and $\vec{W} = \{\ML\}$.  If we set $\ML=0$,
then if $L=0$, $\FF=0$ (so AC2(a) holds) and if $L=1$, then $\FF=1$
(even if $\ML=0$), so AC2(b) and AC2(b$^u$) hold.  However, this
argument required setting $\ML$ to 0, which is not its actual value.
This is not allowed in the modified definition.  According to the modified
definition $L=1 \land \ML =1$ is a cause of $\FF=1$.  Intuitively,
the values of both $L$ and $\ML$ have to change in order to change the
value of $\FF$, so they are both part of a cause, but not causes.
This is but one instance of how parts of causes in the modified HP
definition play a role analogous to causes in the original and updated
HP definitions.
\exam
%\fullv{
It is arguably a feature of the original and modified HP
  definitions that they call $L=1$ and $\ML=1$ causes of
  $\FF=1$, not just 
parts of causes.  \fullv{(But see Example~\ref{xam:fireredux} for more on this
issue.)}  On the other hand, it is arguably a feature of the modified
definition that it can distinguish the causal structure of the
conjunctive and disjunctive cases.
%} 

\xam\label{xam2} Now I consider the rock-throwing example from the
introduction.   The naive causal model would just have endogenous
variables $\BT$, $\ST$, and $\BS$, with the equation $\BS = \ST \lor
\BT$: the bottle shatters if either Suzy or Billy throw.  As observed
in the introduction (and in \cite{HPearl01a}), this naive model does
not distinguish Suzy and Billy, and is isomorphic to the disjunctive
model for the forest fire.  To show that Suzy is the cause, we need a
model that takes into account the reason that we think that Suzy is a
cause, namely, it was her rock that hit the bottle.

As suggested by Halpern and Pearl \citeyear{HPearl01a}, we can
capture this by adding two new variables to the model:
\begin{itemize}
\item $\BH$ for ``Billy's rock hits the (intact) bottle'', with values 0
(it doesn't) and 1 (it does); and
\item $\SH$ for ``Suzy's rock hits the bottle'', again with values 0 and
1.
\end{itemize}
We now modify the equations as follows:
\begin{itemize}
\item $\BS$ is 1 iff one of $\SH$ and $\BH$ is 1;
\item $\SH$ is 1 if $\ST$ is 1; 
\item $\BH = 1$ if $\BT = 1$ and $\SH = 0$.
\end{itemize}
Thus, Billy's 
throw hits if Billy throws {\em and\/} Suzy's rock doesn't hit.
The last equation implicitly  assumes that Suzy throws slightly ahead
of Billy, or slightly harder.  
Call this  model $M_{\RT}$.  
\commentout{
$M_{\RT}$ 
is described by the following graph (where again, the
exogenous variables are omitted).  The asymmetry between
$\BH$ and $\SH$ (in particular, the fact that Billy's throw doesn't
hit the bottle if Suzy throws) is modeled by the fact 
that there is an edge from $\SH$ to $\BH$, but not one in the other
direction; $\BH$ depends (in part) on $\SH$, but not vice versa.
For simplicity, I omit the exogenous variable here.

\vspace{.1in}
\begin{figure}[h]
{\begin{center}
\setlength{\unitlength}{.18in}
\begin{picture}(8,9)
\put(3,0){\circle*{.2}}
\put(0,8){\circle*{.2}}
\put(6,8){\circle*{.2}}
\put(0,4){\circle*{.2}}
\put(6,4){\circle*{.2}}
\put(0,8){\vector(0,-1){4}}
\put(6,8){\vector(0,-1){4}}
\put(0,4){\vector(1,0){6}}
\put(0,4){\vector(3,-4){3}}
\put(6,4){\vector(-3,-4){3}}
\put(3.4,-.2){$\BS$}
\put(-1.7,7.8){$\ST$}
\put(6.3,7.8){$\BT$}
\put(-1.7,3.8){$\SH$}
\put(6.3,3.8){$\BH$}
\end{picture}
\end{center}
}
\vspace{-.1in}
\caption{$M_{\RT'}$---a better model for the rock-throwing example.}\label{fig1}
\end{figure}
}

Taking $u$ to be the context where Billy and Suzy both throw, 
$\ST=1$ of $\BS=1$ in $(M_{\RT},u)$, but $\BT = 1$ is not, according
to all the definitions.  But the arguments are somewhat different.
I start with the argument for the original and updated HP definitions.
To see that $\ST=1$ is a cause according to these definitions, note
that, as usual, it is immediate 
that AC1 and AC3 hold.  For AC2, choose $\vec{Z} = \{\ST,\SH,\BH,\BS\}$,
$\vec{W}=\{\BT\}$,  and $w=0$. 
When $\BT$ is set to 0, $\BS$ tracks $\ST$: if Suzy
throws, the bottle shatters  and if she doesn't throw, the bottle does
not shatter.  To see that $\BT=1$ is \emph{not} a cause
of $\BS=1$, we must check that there is no
partition $\vec{Z} \cup \vec{W}$ of the endogenous variables that
satisfies AC2.
Attempting the symmetric choice with $\vec{Z} = \{\BT,\BH,\SH\,\BS\}$,
$\vec{W}=\{\ST\}$, and $w=0$ 
violates AC2(b) and AC2(b$^u$). To see this, take $\vec{Z}' = \{\BH\}$.  In the
context where Suzy 
and Billy both throw, $\BH=0$.  If $\BH$ is set to 0, the bottle does
not shatter if Billy throws and Suzy does not.  
It is precisely because, in this context, Suzy's throw hits the bottle
and Billy's does not that the original and updated HP definitions
declare Suzy's throw to be the cause of the 
bottle shattering.  AC2(b) and AC2(b$^u$) capture that intuition by
forcing us to consider the contingency where $\BH=0$ (i.e., where $\BH$
takes on its actual value), despite the fact that Billy throws.   
(To show that Billy's throw is not a cause, we also have to check all
the other partitions of the variables; this is left to the reader.)

The modified definition works differently.  First, to show that $\ST=1$ is
cause, we take $\vec{W} = \{\BH\}$ and $w = 0$; that is, we hold $\BH$
at its actual value of 0.  Now if $\ST=0$, then $\BS=0$, showing that
AC2(a$^m$) holds; even if $\BT=1$, the fact that $\BH=0$ means that the
bottle does not shatter.  (Note that we could have also taken $\vec{W} =
\{\BH\}$ in the original and updated definitions to show that $\ST=1$
is a cause of $\BS=1$.)  Showing that Billy's throw is not a cause
is much easier under the modified definition: there are no variables that
can be held at their current value such that if $\BT=0$ we would have
$\BS=0$. Since, in the actual situation, $\ST = \SH = 1$, the bottle
shatters no matter what Billy does.%
\fullv{\footnote{The model $M_{\RT'}$ seems to ``bake in'' the temporal
ordering of events, in particular, that Suzy's rock hits before
Billy's rock.  It is not necessary to do this.
We can allow who hits first to be determined by the context, so that
there may be a context $u'$ where Billy hits first.  This does not
affect the analysis at all.  An alternative approach to incorporating
temporal information is to have time-indexed variables (e.g., to have a
family of variables $\BS_k$ for ``bottle shatters at time $k$'').
In addition to the model used above, 
Halpern and Pearl \citeyear{HP01b} consider a model with time-indexed
variables.  Nothing essential changes in the analysis if we consider such
a model.}}
\exam

I next consider the Hopkins and Pearl \citeyear{HopkinsP02} example
that resulted in the change from the original definition to the
updated definition.

\xam\label{xam:HopkinsP} Suppose that a prisoner dies 
either if $A$ loads $B$'s gun and $B$ shoots, or if $C$ loads and shoots
his gun.  Taking $D$ to represent the prisoner's death and making the
obvious assumptions about the meaning of the variables, we have that
$D= (A\land B) \lor C$.  Suppose that in the actual
context $u$, $A$ loads $B$'s gun, $B$ does not shoot, but $C$ does load
and shoot his gun, so that the prisoner dies.  That is, $A=1$, $B=0$,
and $C=1$. Clearly $C=1$ is a cause of $D=1$. 
We would not want to say that $A=1$ is a cause of $D=1$,
given that $B$ did not shoot (i.e., given that $B=0$).  However,
the original HP definition does exactly that. 
Let $\vec{W} = \{B,C\}$ and consider the contingency where $B=1$ and $C=0$.
It is easy to check that AC2(a) and AC2(b) hold for this contingency,
so under the original HP definition, $A=1$ is a cause of $D=1$.  However,
AC2(b$^u$) fails in this case, since $(M,u) \sat [A \gets 1, C \gets 0](D=0)$.
The key point is that AC2(b$^u$) says that for $A=1$ to be a cause of
$D=1$, it must 
be the case that $D=1$ even if only some of the values in $\vec{W}$ are set
to their values in $\vec{w}$.  In this case, by
setting only $A$ to 1 and leaving $B$ unset, $B$ takes on its original
value of 0, in which case $D=0$.  AC2(b) does not consider this case.

The modified definition also gives the appropriate answer here, but the
argument is simpler.  Clearly $C=1$ is a but-for cause; it is a cause
under the modified definition taking $\vec{W} = \emptyset$.  $A=1$ is not a
cause, since there are no variables whose values we can hold fixed
such that then setting $A=0$ results in $D=0$.  
\exam

Next, consider 
``bogus prevention'' example due to
Hitchcock \citeyear{Hitchcock07} (based on an example due
to Hiddleston \citeyear{Hiddleston05}), which motivated the
addition of normality considerations to the HP definition
\shortv{\cite{HH11}.}{\cite{Hal39,HH11}.}  

\xam\label{xam:bogus}
Assassin is in possession of a lethal poison, but has a last-minute
change of heart and refrains from putting it in Victim's coffee.
Bodyguard puts antidote in the coffee, which would have neutralized the
poison had 
there been any.  Victim drinks the coffee and survives.  
Is Bodyguard's putting in the antidote a cause of Victim surviving?
Most people would say no, but
according to the original and updated HP definition, it is.  For in
the contingency where Assassin puts in the poison, Victim survives iff
Bodyguard puts in the antidote.   However, according to the modified
definition, it is not.  Even if Bodyguard doesn't put in the antidote,
Victim survives, as long as we hold any subset of the other variables
at their actual values. 

Bodyguard putting in the antidote is part of a cause under the
modified definition. Bodyguard putting in antidote and 
Assassin not putting in poison together form a cause.  This does not
seem so unreasonable.  If Assassin had poisoned the coffee and
Bodyguard hadn't put in antidote, the king would have died.
However, intuitions may differ here.  We might argue that we don't
need a cause for an event that was expected all along.  Here normality
considerations can help.  If we use the extension of the HP
definitions to deal with normality proposed by Hitchcock and Halpern
\citeyear{HH11} (which applies without change to the modified
definition), then under reasonable assumptions, the witness to Bodyguard
putting in antidote being a cause of Victim surviving is the world
where Bodyguard 
doesn't put in antidote and Assassin puts in poison.  This world is
not at least as 
normal as the actual world (arguably, it is incomparable in normality
to the actual world), so the Halpern and Hitchcock approach would not
declare Bodyguard (part of) a cause, according to any variant of the
HP definition. \exam

%This example can also be dealt with by adding extra
%  variables or appealing to normality considerations \cite{HH11}, but
%  such moves are not necessary with the modified definition.
%\exam

Arguments similar to those used in Example~\ref{xam:bogus} also 
show that the modified definition gives the 
  appropriate answer in the case of Hall's \citeyear{Hall07}
  \emph{nonexistent threat}.  Here $C=1$ would have prevented $E=1$
  had $B$ been 1, but in the actual context, $B=0$ (so we can view $B$
  as a potential threat which is nonexistent in the actual context,
  since $B=0$).  The original and updated HP definitions declare $C=1$
  to be a cause, contrary to intuition (by considering the contingency
  where $B=1$); the modified HP definition does not.

Halpern \citeyear{Hal44} discussed a number of examples from the
literature purportedly showing problems with the updated definition,
and shows that they can be dealt with by using what is arguably a
better model of the situation, with extra variables.  
These problems can be dealt with by the modified definition, without
introducing extra variables.  \shortv{I give one example here, due to 
Weslake \citeyear{Weslake11}.}
\fullv{I illustrate this with the following example, 
due to Weslake \citeyear{Weslake11}.}

\xam\label{xam:Weslake}  A lamp $L$ is controlled by three
switches, $A$, $B$, and $C$, each of which has three possible positions,
$-1$, $0$, and $1$.  The lamp switches on iff two or more of the
switches are in same position.  Thus, $L=1$ iff $
(A=B) \lor (B=C) \lor (A=C).$
Suppose that, in
the actual context, $A=1$, $B=-1$, and $C=-1$.  Intuition
suggests that while $B=-1$ and $C=-1$ should be causes of $L=1$, $A=1$
should not be; since the setting of $A$ does not match that of either $B$
or $C$, it has no causal impact on the outcome.
The original and updated HP definitions indeed declare $B=-1$ and $C=-1$ to be
causes; unfortunately, they also declare $A=1$ to be a cause.  For in the
contingency where $B=1$ and $C=-1$, if $A=1$ then $L=1$, while if $A=0$ then
$L=0$.  The modified definition declares $B=-1$ and $C=-1$ to be
causes (again, these are but-for causes, so all the definitions
agree), but it does \emph{not} declare $A=1$ to be a cause.  The
contingency where $B=1$ and $C=-1$ cannot be considered by the
modified definition.
\shortv{\exam}

\fullv{
Example~\ref{xam:Weslake} is dealt with in \cite{Hal44} by considering two
stories for why the lamp goes on: the first is Weslake's story (it
switches on if at least two of $A$, $B$, and $C$ have the same
setting); the second takes the lamp to switch if there is a setting $i$
(either $-1$, $0$, or $1$) such that none of the lamps have setting
$i$.  Both stories are described by the same equation for $L$.  But in
the second story, it seems reasonable to call $A=1$ a cause of $L=1$.
By adding variables to the model, we can distinguish these
stories; in these richer models, the original and updated HP
definitions make the ``right'' causal judgments.  The modified definition
agrees with these judgments.  I think that there are good reasons for
considering the richer models.  Indeed, if we start with the intuition
given by the second story, then under the modified definition, it is
necessary to consider the richer model to declare $A=1$ a cause.
Roughly speaking, this
is because, under the modified definition, there must be some variable
whose value in the real world demonstrates the causality.  The
simple model whose variables are only $A$, $B$, $C$, and $L$ is not
rich enough to do this.  

Halpern \citeyear{Hal44} also considers an example of Spohn
\citeyear{Spohn12} which is similar in spirit.  Again, the modified
definition handles it appropriately, without needing to add variables to
the model.
\exam  

Example~\ref{xam:Weslake} (as well as
Example~\ref{xam:glymour} below and other examples considered by
Halpern \citeyear{Hal44}) show that by adding variables to describe
the mechanism of causality, we can distinguish two situations which
otherwise seem identical.  As the following example (suggested by an anonymous
reviewer of the paper) shows, adding variables that describe the
mechanism also allows us to convert a part of a cause according to the
modified HP definition to a cause.

\xam\label{xam:fireredux}  Suppose that we add variables $A$, $B$, and
$C$ to the disjunctive forest-fire example (Example~\ref{xam:ff}),
where $A=L \land 
\neg \ML$, $B = \neg L \land \ML$, and $C=L \land \ML$.  We then
replace the earlier equation 
for $\FF$ (i.e., $\FF = L \lor \ML$)
by $\FF = A \lor B \lor C$.  The variables $A$, $B$, and $C$ can be
viewed as describing the mechanism by which the forest fire happened.
Did it happen because of the dropped match only, because of the
lightning only, or because of both?  Alternatively, $A$, $B$, and $C$
could describe the intensity of the forest fire (it might be more
intense if both the arsonist drops a match and the lightning strikes).
Whatever the interpretation, in this model, not only are $L=1$ and
$\ML=1$ causes of $\FF=1$ according to the original and updated HP
definitions, they are also causes according to the modified
definition.  For if we fix $A$ and $B$ at their actual values of 0,
then $\FF=0$ if $L$ is set to 0, so AC2(a$^m$) is satisfied and $L=1$
is a cause; an analogous argument applies to $\ML$.  

I would argue that this is a feature of the modified definition, not a
bug.  Suppose, for 
example, that we interpret $A$, $B$, and $C$ as describing the
mechanism by which the fire occurred.  If these variables are in the
model, then that suggests that we care about the mechanism.  The fact
that $L=1$ is part of the reason that $\FF=1$ occurred thanks to mechanism $C$.
While the forest fire would still have occurred if the lightning
hadn't struck, it would have due to a different mechanism.  The same
argument applies if we interpret $A$, $B$, and $C$ as describing the
intensity of the fire (or any other feature that differs depending on
whether there was lightning, a dropped match, or both).

In the original model, we essentially do not care about the details of
how the fire comes about.  Now suppose that we care only about whether
lightning was a cause.  In that case, we would add only the variable $B$,
with $B = \neg L \land \ML$, as above, and set $\FF = L \lor B$.
In this case, in the context where $L = \ML = 1$, all three variants
of the HP definition agree that only $L=1$ is a
cause of $\FF=1$; $\ML=1$ is not (and is not even part of a cause).
Again, I would argue that
this is a feature.  The structure of the model tells us that we should
care about how the fire came about, but only to the extent of whether
it was due to $L=1$.  In the actual context, $\ML=1$ has no impact on
whether $L=1$.
\exam

The next example, due to Glymour et al. \citeyear{Glymouretal10}, is
also discussed by Halpern \citeyear{Hal44}.

\xam\label{xam:glymour}
A ranch has five individuals: $a_1, \ldots, a_5$.  They
have to vote on two possible outcomes: staying around the campfire
($O=0$) or going on a round-up ($O=1$).  Let $A_i$ be the 
variable denoting $a_i$'s vote, so $A_i = j$ if $a_i$ votes for outcome
$j$.  There is a complicated rule for deciding on the outcome.  If $a_1$
and $a_2$ agree (i.e., if $A_1 = A_2$), then that is the outcome.  If
$a_2, \ldots, a_5$ agree, and $a_1$ votes differently, then then outcome
is given by $a_1$'s vote (i.e., $O = A_1$).  Otherwise, majority rules.
In the actual situation, $A_1 = A_2 = 1$ and $A_3 = A_4 = A_5 = 0$, so

Using the obvious causal model with just the variables $A_1, \ldots, A_5,
O$, with an equation describing $O$ in terms of $A_1, \ldots,
A_5$, it is almost immediate that $A_1 = 1$ is a cause of $O=1$
according to all three definitions, since it is a but-for cause.
Under the original and updated HP definitions, it is not hard to show that
$A_2 = 1$, $A_3 = 0$, $A_4 =0$, and $A_5 = 0$ are also causes.
For example, to see that $A_2 = 1$
is a cause, consider the contingency where $A_3 = 1$.  Now if $A_2 = 0$,
then $O=0$ (majority rules); if $A_2 = 1$, then $O=1$, since $A_1 = A_2
=1$, and $O=1$ even if $A_3$ is set back to its original value of 0.
However, under the modified definition, only $A_1 = 1$ is a cause.

In this case, my intuition declares both $A_1 = 1$ and $A_2 = 1$
causes.  As suggested in \cite{Hal44}, this outcome can be realized by
adding variables to describe the mechanism that brings about the
result; that is, does $O$ have its value due to the fact that (1) $a_1$
and $a_2$ agreed, (2) $a_1$ was the only one to vote a certain way, or
(3) majority ruled.  
Specifically, we can add three new variables, 
$M_1$, $M_2$, and $M_3$.  These variables have
values in $\{0,1,2\}$, where $M_j = 0$ if mechanism $j$ is active and
suggests an outcome 0, $M_j = 1$ if mechanism $j$ is active and suggests
an outcome of 1, and $M_j = 2$ if mechanism $j$ is not active.  (We
actually don't need the value $M_3 = 2$; mechanism 3 is always active,
because there is always a majority with 5 voters, all of whom must vote.)
Note that at most one of the first two mechanisms can be active.  We
have obvious equations linking the value of $M_1$, $M_2$, and $M_3$ to
the values of $A_1, \ldots, A_5$.  In this model, it is easy to see
that all three definitions agree that $A_1 = 1$ and $A_2=1$ are both
causes of $O=1$.  Intuitively, this is because the second mechanism
was the one that led to the outcome. \exam

\xam As Livengood \citeyear{Liv13} points out, under the original and updated
definitions, if there is a 17--2 vote for candidate $A$ over candidate
$B$, then all of the 17 voters for $A$ are considered causes of $A$'s
victory, and none of the voters for $B$ are causes of the victory.
On the other hand, if we add a third candidate $C$, and the vote
is 17--2--0, then the voters for $B$ suddenly become causes of $A$'s
victory as well.  
To see this, consider a contingency where 8 of the voters for $A$ switch
to $C$.  Then if one of the voters for $B$ votes for $C$, the result is
 a tie;
if that voter switches back to $B$, then $A$ wins (even if some subset
of the voters who switch from $A$ to $C$ switch back to $A$). 
Under the modified definition, any subset of 10 voters for $A$ is a
cause of $A$'s victory, but the voters for $B$ are not causes of $A$'s
victory.  
\exam

The following example is due to Hall \citeyear{Hall01}, and was
discussed by Halpern and Pearl \citeyear{HP01b}:
\xam  
The engineer is standing by a switch in the railroad tracks.  A train
approaches in the distance.  She flips the switch, so that the train
travels down the right-hand track, instead of the left.  Since the
tracks reconverge up ahead, the train arrives at its destination all
the same.

If we model this story using three variables---%
$F$ for ``flip'', with values 0 (the engineer doesn't flip the
switch) and 1 
(she does);
$T$ for ``track'', with values 0 (the train goes on the left-hand
track) and 1 (it goes on the right-hand track); and
$A$ for ``arrival'', with values 0 (the train does not arrive at
the point of reconvergence) and 1 (it does)---
then all three definitions agree that flipping the switch is not a
cause of the train arriving.  
Now, following Halpern and Hitchcock \citeyear{HH10},
 suppose that we replace $T$ with two binary variables, 
$\LB$ (which is 0 if the left-hand track is not blocked, and 1 if it is)
and $\RB$.  We have the obvious equations connecting the variables.
In the actual context $F=1$ and $\LB=\RB=0$.  Under the original and
updated HP definitions, $F=1$ is a cause of $A=1$.
For in the contingency where $\LB=1$, if $F=1$, the train arrives,
while if $F=0$, the train does not arrive.  

Roughly speaking, this was dealt with by 
Halpern and Hitchcock \citeyear{HH10} by observing that the
contingency where $\LB=1$ is abnormal; contingencies that are less
normal than the actual situation are not considered.  However,
Schumacher \citeyear{Schum14} pointed out that this approach runs into
problems when we consider the context where both tracks are blocked.  In
this case, the original and updated HP definitions declare the flip a
cause of the train not arriving  (by considering the contingency where
$\LB=0$).  And now normality considerations don't help, since this
contingency is more normal than the actual situation, where the track
is not blocked.

With the modified definition, this becomes a non-problem.  Flipping
the switch is
not a cause of the train arriving if both tracks are unblocked, nor is it
a cause of the train not arriving of both tracks are blocked.

\newcommand{\LT}{\mbox{{\it LT}}}
\renewcommand{\RT}{\mbox{{\it RT}}}
Hall's \citeyear{Hall07} model of the story uses different variables.
Essentially, instead of the variables $\LB$ and $\RB$, he has
variables $\LT$ and $\RT$, for ``train went on the left track'' and
``train went on the right track''.  In the actual world, $F=1$,
$\RT=1$, $\LT=0$, and $A=1$.  Now $F=1$ is a cause of $A=1$, according
to the modified definition (as well as the original and updated
HP definitions).  If we simply fix $\LT=0$ and
set $F=0$, then $A=0$.  But here normality conditions do apply: the
world where the train does not go on the left track despite the switch
being set to the left is less normal than the actual world.
\exam

The final two examples consider cases where the the modified
definition by itself arguably does not give the appropriate answer, but it does
when combined with considerations of normality (in the first example)
and responsibility and blame (in the second example).  
The first of these examples is taken from Hitchcock
\citeyear{Hitchcock07}, where 
it is called ``counterexample to Hitchcock''.  Its structure is
similar to Hall's \emph{short-circuit} example
\citeyear{Hall07}[Section 5.3]; the same analysis applies to
both.
 
\xam\label{xam:circuit} Consider a variant of the bogus prevention
problem.
Again, Bodyguard puts an antidote in Victim's coffee, but now Assassin
puts the poison in the coffee.  However, Assassin would not
have put the poison in the coffee if Bodyguard hadn't put the antidote
in.  (Perhaps Assassin is putting in the poison only to make Bodyguard
look good.)   Now Victim drinks the coffee and survives.  

Is Bodyguard putting in the antidote a cause of Victim surviving?
It is easy to see that, according to all three variants of the
definition, it is. 
If we fix Assassin's action, then Victim survives if and only if
Bodyguard puts in the antidote.   Intuition suggests that this is
unreasonable.  By putting in the antidote, Bodyguard neutralizes the 
effect of the other causal path he sets in action: Assassin putting
in the poison.  

Although no variant of the HP definition can deal with this example,
as already pointed out by Hall \citeyear{Hall07} and Hitchcock
\citeyear{Hitchcock07}, by taking into account normality considerations, we can
recover our intuitions.  Using, for example, the extension of the HP
definitions to deal with normality proposed by Hitchcock and Halpern
\citeyear{HH11}, the witness to Bodyguard putting in the
antidote being a cause of Victim surviving is the world where Bodyguard
doesn't put in the antidote but Assassin puts in the poison anyway,
directly contradicting the story.  This is arguably an abnormal world
(much less normal than the actual world), and thus should not be
considered when determining causality, according to the
Halpern-Hitchcock approach (and, for much the same reasons, should not
be considered a cause in the models proposed by Hall
\citeyear{Hall07} and Hitchcock \citeyear{Hitchcock07}).
\exam
}

The final example touches on issues of legal responsibility.
%in the   law.
\xam
Suppose that two companies both dump pollutant into the
river.  Company $A$ dumps 100 kilograms of pollutant; company $B$ dumps
60 kilograms.  This causes the fish to die.  Biologists determine that $k$
kilograms of pollutant sufficed to cause the fish to die.  Which company is
the cause of the fish dying if $k=120$, if $k=80$, and if $k=50$?

It is easy to see that if $k=120$, then both companies are causes of
the fish dying, according to all three definitions (each company is a
but-for cause of the outcome).  If $k=50$, then each company is still
a cause according to the original and updated HP definitions.  For
example, to see that company $B$ is a cause, we consider the
contingency where company $A$ does not dump any pollutant.  Then the
fish die if company $B$ pollutes, but survive if $B$ does not pollute.
With the modified definition, neither company individually is a cause;
there is no variable
that we can hold at its actual value that would make company $A$ or
company $B$ a but-for cause.    However, both companies together are the cause.

The situation gets more interesting if $k=80$.  Now the modified 
definition says that only $A$ is a cause; whether or not we keep $A$
fixed at dumping 100 kilograms of pollutant, what $B$ does has no impact.
The original and updated definitions also agree that $A$ is a cause if
$k=80$.  
Whether $B$ is a cause depends on the possible amounts of pollutant
that $A$ can dump.  If $A$ can dump only 0 or 100 kilograms of pollutant,
then $B$ is not a cause; no setting of $A$'s action can result in
$B$'s action making a difference.  However, if $A$ can dump some
amount between 21 and 79 kilograms, then $B$ is a cause.

It's not clear what the ``right'' answer should be here if $k=80$.  The law
typically wants to declare $B$ a contributing cause to the death of the
fish (in addition to $A$), but should this depend on the amount of
pollutant that $A$ can dump?
This issue is perhaps best dealt with by considering an extension to
the HP approach that takes into account 
\emph{degree of responsibility} and \emph{degree of blame}
\cite{ChocklerH03,ZGL12}.  Degree of blame, in particular, takes into
account the agent's uncertainty about how much pollutant was dumped.
Under reasonable assumptions about the agent's degree of uncertainty
regarding how likely various amounts of
pollutant are to be dumped, $B$ will get some degree of blame under
the modified definition, even when it is not a cause.
\exam

\fullv{
\section{Comparison to other approaches}\label{sec:compare}
The key difference between the modified HP definition on the one hand
and the original and updated HP definitions on the other is the
insistence that the contingency considered in AC2(a) be one where all
the variables take their initial values.  Doing so makes it clear that
the sufficient 
condition (AC2(b)/AC2(b$^u$)) is needed only to handle cases where the
variables in the contingency considered take on non-actual values.  
The idea of keeping
variables fixed at their actual value when considering changes also arises
in other definitions of causality.  I focus on three of them here: Pearl's
\citeyear{pearl:98c,pearl:2k} \emph{causal beam} definition, what Hall
\citeyear{Hall07} calls the \emph{H-account}, and Hitchcock's
\citeyear{hitchcock:99} definition of actual causality.  I briefly
compare these alternatives here to the modified HP definition here.

All the variants of the HP definition were inspired by Pearl's
original notion of a \emph{causal beam} \cite{pearl:98c}.%
\footnote{The definition of causal beam in \cite{pearl:2k}[Chapter 10]
  is a modification of the original definition that takes into account
  concerns raised in an early version of \cite{HPearl01a}.  The
  differences are not relevant to this discussion.}
It would take us too far afield to go into the details of the causal
beam definition here.  The definition was abandoned due to 
problems. (See Example~\ref{xam:voting} below.)  However, it is worth
noting  that, roughly speaking, according to this definition, $A$ only
qualifies as an actual cause of $B$ if something like AC2(a$^m$)
rather than AC2(a) holds; otherwise it is called a \emph{contributory
  cause}.  The distinction between actual cause and contributory cause
is lost in the original and updated HP definition.  To some extent, it
resurfaces in the modified HP definition, since in some cases what the
causal beam definition would classify as a contributory cause but not an actual
cause would be classified as part of a cause but not a cause according to the
modified HP definition.

Hall~\citeyear{Hall07} considers a variant of the HP definition that
he calls the \emph{H-account}.  This variant, as well as Hitchcock's
definition, involve causal paths.  A \emph{causal path} from $X$ to
$Y$ in $(M,\vec{u})$ is a sequence $(Z_0, \ldots, Z_k)$ of variables
such that $X=Z_0$, $Y=Z_k$, and $Z_{i+1}$ depends on $Z_i$ 
(i.e., if there is some setting of all the variables in
$\U \union \V$ other than $Z_{i+1}$ and $Z_i$ such that varying the value of
$Z_{i}$ in the equation $F_{Z_{i+1}}$ for $Z_{i+1}$ changes the value
of $Z_{i+2}$).  Hall takes $X=x$ to be a cause of $Y=y$ according to the
H-account in context $(M,\vec{u})$ if there is a causal path from $X$
to $Y$, some setting $\vec{w}$ of variables $\vec{W}$  not on this
causal path and setting $x'$ of $X$ such that AC2(a) holds, and for
all variables $Z$ on the causal path, $(M,\vec{u}) \sat [\vec{W} \gets
  \vec{w}](Z=z)$, where $z$ is the actual value of $Z$ in
$(M,\vec{u})$ (i.e., $(M,\vec{u}) \sat Z = z$).  This is clearly a
strengthening of AC2(b); if $X=x$ is a cause of $Y=y$ in $(M,\vec{u})$
according to the H-account, then it is clearly a cause according to
the original and updated HP definitions.

Unfortunately, the H-account is too strong, as the following example
(taken from \cite{HP01b}) shows:

\xam\label{xam:voting}  Suppose that two people vote for a measure, which
will pass if at least one of them votes in 
favor.  In fact, both of them vote in favor, and the measure passes.
This is isomorphic to the disjunctive version of the forest-fire
example, but there is a twist: there is a voting machine 
that tabulates the votes.  Thus, the model has four exogenous
variables: $V_1$, $V_2$, $M$, and $P$.  $V_i$ represents voter $i$'s
vote, $M = V_1 + V_2$ (so $M$ can have values in $\{0,1,2\}$) and 
$P=1$ (the measure passes) if and only if  $M \ge 1$.  
In this model, it is easy to see that $V_1=1$ and $V_2=1$ are causes of
$M$ according to the original and updated HP definitions, and parts of
causes according to the modified HP definition (which calls $V_1 = 1
\land V_2 = 1$ a cause).  However, neither $V_1=1$ nor $V_2=1$ is a
cause according to the H-account.  For example, to show
that $V_1=1$ is a cause, we would need to set $V_2=0$.  But the causal
path from $V_1$ to $P$ must go through $M$ (just changing $V_1$ while
keeping $M$ fixed has no effect on $P$), and if $V_2=0$, $M$ does not
have it original value. As pointed out by Halpern and Pearl
\citeyear{HP01b}, this example also causes problems for the causal
beam definition; $V_1=1$ is neither an actual nor a contributory cause
of $P=1$ according to the causal beam definition.  In general, in
showing that $X=x$ is a cause of $Y=y$, it seems to be asking too much
to require that changes in the
off-path variables have no effect on variables along the causal path;
it seems to suffice to require that changes in the off-path variables
not affect the final outcome $Y=y$, 
\exam

I conclude this section by considering 
the definition of actual causality proposed by Hitchcock
\citeyear{hitchcock:99}, which is perhaps closest in spirit the
modified HP definition.
Given a causal path $P$ from $X$ to $Y$, $M^P$ is the \emph{reduction of
  $M$ along $P$} if 
$M^P$ obtained from $M$ by replacing the equation 
for each variable $W$ not on the path by the equation
$W=w$, where $w$ is such that $(M,\vec{u}) \sat W=w$.%
\footnote{Hitchcock does this replacement only for variables $W$ that
lie   on some path from $X$ to $Y$.  Doing the replacement for all
off-path variables has no affect on Hitchcock's definition.}
Hitchcock takes $X=x$ to be a cause of 
$X=x$ if there is a path $P$ from $X$ to $Y$ such that $X=x$
is a but-for cause of $Y=y$ in $M^P$.
Hitchcock's insistence on looking at a single 
causal path causes problems, as the following example shows.

\xam\label{xam:hitchcock} Consider a model $M$ with four binary
endogenous variables, $A$, $B$, 
$C$, and $D$.  The value of $A$ is set by the context; we have the
equations $B = A$, $C=A$, and $D = B \lor C$.  In the actual context
$A=1$, so $B=C=D=1$.  $A=1$ is a but-for cause of $D=1$, so it is a
cause according to all three variants of the HP definition.  There are
two causal paths from $A$ to $D$: $P_1 = (A,B,D)$ and $P_2 = (A,C,D)$.
But $A=1$ is not a but-for cause of $D=1$ in either $M^{P_1}$ or $M_{P_2}$.
For example, in the case of $M_{P_1}$, we must fix $C$ at 1, so $D=1$,
independent of the value of $A$.  There does not seem to be an obvious
change to Hitchcock's definition that would deal with this problem and
maintain the spirit of the modified HP definition.%
\footnote{Hitchcock also
considers a variant of his definition where he allows the variables $W$
off the path to change values to within what he calls their \emph{redundancy
  range}.  This change will deal with the problem in this example, but
the resulting definition is then no longer in the spirit of the
modified definition.  It is somewhat closer to the original HP
definition, and suffers from other problems.}
\exam
}

\section{The complexity of determining causality}\label{sec:complexity}
The complexity of determining causality for the original and updated
HP definitions has been completely characterized.  To explain the
results, I briefly review some complexity classes:

Recall that the \emph{polynomial hierarchy} is a hierarchy of complexity
classes that generalize $\NP$ and co-$\NP$.  Let $\Sigma^P_1 =
\NP$ and $\Pi^P_1 = \mbox{co-}\NP$.  For $i > 1$, define  $\Sigma^P_i =
\NP^{\Sigma_{i-1}^P}$ and $\Pi_i^P = (\mbox{co-}\NP)^{\Sigma_{i-1}^P}$, where, in
general, $X^Y$ denotes the class of problems solvable by a Turing
machine in class $X$ augmented with an oracle for a problem complete for
class $Y$ \cite{Stock}.  The classes 
$D^P_k$ were defined by Aleksandrowicz et al. \citeyear{ACHI14} as
follows.
% \begin{definition}\label{def-dk}
For  $k = 1, 2, \ldots$,
\[D^P_k = \{ \Lan: \exists \Lan_1, \Lan_2: \Lan_1 \in \Sigma^P_k, \Lan_2
\in \Pi^P_k,  
\Lan = \Lan_1 \cap \Lan_2 \}. \]
%\end{definition}
The class $D^P_1$ is the well-known complexity class $\DP$
\cite{PY}. It contains
\emph{exact} problems such as the language of pairs $\zug{G,k}$, 
where $G$ is a graph that has a maximal clique of size exactly $k$.
As usual, a language $\Lan$ is \emph{$D^P_k$-complete} if it is in
$D^P_k$ and is the ``hardest'' language in $D^P_k$, in the sense that
there is a polynomial time reduction from any language $\Lan' \in D^P_k$
to $\Lan$. 

As shown by Eiter and
Lukasiewicz \citeyear{EL01} and Hopkins \citeyear{Hopkins01}, under the
original HP definition, we can always take causes to be single
conjuncts.  Using this fact, Eiter and Lukasiewicz showed that, under
the original HP definition, the
complexity of determining whether $X=x$ is a cause of $\phi$ is
$\Sigma_2^P$-complete.
%and $\NP$-complete if we restrict 
%all variables to being binary (i.e., taking on only two possible
%values).  
Halpern \citeyear{Hal39} showed that for the updated
definition, we cannot always take causes to be single conjuncts;
Aleksandrowicz et al. \citeyear{ACHI14} showed that the complexity of
computing whether $\vec{X} = \vec{x}$ is a cause of $\phi$ under the
updated HP definition is $D_2^P$-complete.  Roughly speaking, this is
because, under the updated HP definition, checking AC2 is
$\Sigma_2^P$-complete and checking AC3 is $\Pi_2^P$-complete.  With
the original HP definition, checking AC3 is vacuous, because causes
are always single conjuncts.  
%With the modified HP definition, the check becomes nontrivial.

I show here that with the modified definition, the complexity of causality
is $D_1^P$-complete; checking AC2 drops from $\Sigma_2^P$ to $\NP$,
while checking AC3 drops from $\Pi_2^P$ to co-$\NP$.

\thm\label{thm:complexity} The complexity of determining whether $\vec{X} = \vec{x}$ is a cause
of $\phi$ in $(M,\vec{u})$ is $D_1^P$-complete.
\ethm
\prf
The argument is similar in spirit to that of \cite{ACHI14}.  Formally,
we want to show that the language $\Lan =
 \{ \pair{M,\vec{u},\phi, \vec{X}, \vec{x}} :
 (\vec{X}=\vec{x}) \mbox{ satisfies 
AC1, AC2(a$^m$), and AC3 for $\phi$ in  $(M,\vec{u})$} \}$ is $D_1^P$-complete.
Let 
$$\begin{array}{lll}
 \Lsingle = &\{ \pair{M,\vec{u},\phi, \vec{X}, \vec{x}} :
 (\vec{X}=\vec{x}) \mbox{ satisfies} \\
&\ \ \mbox{AC1 and AC2(a$^m$) for $\phi$ in  $(M,\vec{u})$} \}, \\
 \Lmin = &\{ \pair{M,\vec{u},\phi, \vec{X}, \vec{x}} :
 (\vec{X}=\vec{x}) \mbox{ satisfies}\\
&\ \  \mbox{AC1 and AC3 for  $\phi$ in
   $(M,\vec{u})$} \}. 
\end{array}
$$
Clearly $\Lan = \Lsingle \inter \Lmin$.  Thus, it suffices to show that
$\Lsingle$ is $\NP$-complete and $\Lmin$ is co-$\NP$-complete.  
\shortv{I defer details to the full paper. \eprf}

\fullv{
It is easy to see that $\Lsingle$ is in $\NP$.  Checking that AC1
holds can  
be done in polynomial time, and to check whether AC2(a$^m$) holds, we
can guess $\vec{W}$ and $\vec{x}'$, and check in polynomial time that
$(M,\vec{u}) \sat [\vec{X} \gets \vec{x}', \vec{W} \gets \vec{w}] \neg
\phi$ (where $\vec{w}$ is such that $(M,\vec{u}) \sat \vec{W} =
\vec{w}$).  Similarly, $\Lmin$ is in co-$\NP$, since checking whether
AC3 is not satisfied can be done by guessing a counterexample and
verifying.

To see that $\Lsingle$ is $\NP$-hard, we reduce propositional
satisfiability to $\Lsingle$. 
Given an arbitrary formula $\phi$ with primitive propositions $X_1,
\ldots, X_n$, consider the causal model $M$ with endogenous variables
$X_0, \ldots, X_n, Y$, one exogenous variable $U$, equations $X_0 = U$,
$X_i = X_0$ for $i = 1, \ldots, n$ and $Y = X_0 \land \phi$.  Clearly,
$(M,0) \sat X = 0 \land Y=0$.
Thus, $X = 0$ satisfies AC1 and AC2(a$^m$) for $Y=0$
in $(M,0)$ exactly if there is some subset $\vec{W}$ of $\{X_0,
\ldots, X_n\}$ such that holding the variables in $\vec{W}$ fixed at 0
and setting all the remaining variables to 1 results in $Y = 1$.  But
in such an assignment, we must 
have $X_0 = 1$; the setting of the remaining variables gives a
satisfying assignment for~$\phi$.

To see that $\Lmin$ is co-$\NP$-hard, we reduce unsatisfiability to
$\Lmin$.  The idea is very similar to that above.  Suppose we want to
check if $\phi$ is unsatisfiable.  
We now use endogenous variables $X_0, \ldots, X_n,
X_{n+1}, Y$.  We still have the equations $X_i = U$ for $i = 0,
\ldots, n+1$, but now the equation for $Y$ is $Y = X_0 \land \phi \land \neg
X_{n+1}$.  Call this model $M'$.
Again we have $(M',0) \sat \vec{X} = \vec{0} \land Y=0$.
It is easy to see that 
$\vec{X} = \vec{0}$ satisfies AC1 and AC3 for $Y=0$
in $(M,0)$ exactly if $\phi$ is unsatisfiable.

This completes the proof.
\eprf

Things simplify if we restrict to causes that are single conjuncts,
since in that case, AC3 holds vacuously.

\thm The complexity of determining whether $X = x$ is a cause
of $\phi$ in $(M,\vec{u})$ is $\NP$-complete.
\ethm

\prf The proof follows almost immediately from the proof of
Theorem~\ref{thm:complexity}.  Now
we want to show that $\Lan' =
 \{ \pair{M,\vec{u},\phi, \vec{X}, \vec{x}} :
 (X=x) \mbox{ satisfies 
AC1, AC2(a$^m$), and AC3 for $\phi$ in  $(M,\vec{u})$} \}$ is {\sl
   NP}-complete. AC3 trivially holds and, as we have observed,
 checking that AC1 and AC2(a$^m$) holds is in $\NP$.  Moreover, the
 proof of Theorem~\ref{thm:complexity} 
 shows that AC2(a$^m$) is {\sl NP}-hard even if we consider
 only singleton causes.
\eprf
}

\section{Conclusion}\label{sec:conclusion}
The modified HP definition is only
a relatively small modification of the original and updated HP
definitions (and, for that matter, of other definitions that have been
proposed).  But the modification makes it much simpler (both
conceptually and in terms of its complexity).  Moreover,
as the example and discussion 
\fullv{in Sections~\ref{sec:compare} and}
Section~\ref{sec:examples} show, small changes can have
significant effects.  I have shown that the modified HP definition
does quite well on many of the standard counterexamples in the
literature.  (It also does well on many others not discussed in
the paper.)   When combined appropriately with notions of
normality and responsibility and blame, it does even better.
Of course, this certainly does not prove that the modified HP
definition is the ``right'' definition.  
The literature is littered with attempts to define actual causality
and counterexamples to them.    
%Someone may 
%come up tomorrow with a devastating example on which it does badly.
This suggests that we should keep trying to understand the
space of examples, and how causality interacts with normality,
responsibility, and blame.

\fullv{
\newenvironment{RETHM}[2]{\trivlist \item[\hskip 10pt\hskip\labelsep{\sc #1\hskip 5pt\relax\ref{#2}.}]\it}{\endtrivlist}
\newcommand{\rethm}[1]{\begin{RETHM}{Theorem}{#1}}
\newcommand{\erethm}{\end{RETHM}}

\appendix

\section{Proof of Theorem~\ref{thm:connections}}
In this appendix, I prove Theorem~\ref{thm:connections}.    I repeat
the statement of the theorem for the reader's convenience.

\rethm{thm:connections}
If $X=x$ is part of a cause of $\phi$ in $(M,\vec{u})$
  according to the modified HP definition, then 
$X=x$ is a cause of $\phi$ in $(M,\vec{u})$
  according to both the original and updated HP definitions. 
\erethm

\prf
Suppose that $X=x$ is part of a cause of $\phi$ in
$(M,\vec{u})$ according to the modified HP definition, so that there
is a cause 
$\vec{X} = \vec{x}$ such that $X = x$ is one of its conjuncts.  I
claim that $X=x$ is a cause of $\phi$ in $(M,\vec{u})$ according to
the original HP 
definition.   
By definition, there must exist a value $\vec{x}' \in
\R(\vec{X})$ and a set $\vec{W} \subseteq 
\V - \vec{X}$ such that if $(M,\vec{u}) \sat \vec{W} = \vec{w}$, then
$(M,\vec{u}) \sat [\vec{X} \gets \vec{x}', \vec{W} \gets
    \vec{w}]\neg \phi$.  Moreover, $\vec{X}$ is minimal.
%Clearly this is true if $|\vec{X}| = 1$.  In that case, $\vec{X} = X$.
%AC1, AC2(a), and AC2(b) 
%hold for $X=x$ (taking $\vec{Z} = \V - \vec{W}$), and AC3 trivially holds 
%Since $\vec{X}$ is a singleton.

To show that $X=x$ is a cause according to the original HP definition,
we must find an appropriate witness.  If $\vec{X} = \{X\}$, then it is
immediate that $(\vec{W},\vec{w},x')$ is a witness.  
If $|\vec{X}| > 1$, suppose without loss of generality that $\vec{X} =
\<X_1, \ldots, X_n\>$, and $X = X_1$.  In general, if $\vec{Y}$ is a
vector, I write $\vec{Y}_{-1}$ to denote all components of the vector
except the first one, so that $\vec{X}_{-1} = \<X_2,
\ldots, X_n\>$.
%and let $\vec{x}^*$ consist of those
%components of $\vec{x}$ correspond to $\vec{X}^*$, 
I want to show that $X_1 = x_1$ is a cause of $\phi$ in
$(M,\vec{u})$ according to the original HP definition.  Clearly,
$(M,\vec{u}) \sat X_1 = x_1 \land \phi$, since 
$\vec{X} = \vec{x}$ is a cause of $\phi$ in $(M,\vec{u})$ according to
the modified HP definition, so AC1 holds.  
%Let $\vec{W}' =
%\vec{X}_{-1} \cdot \vec{W} $ and let $\vec{w}' = \vec{x}'_{-1}
%\cdot \vec{w}$, 
The obvious candidate for a witness for AC2(a) is 
$(\vec{X}_{-1}\cdot \vec{W},\vec{x}_{-1}'\vec{w}, x_1')$,
where $\cdot$ is the operator that concatenates two
vectors.    
This satisfies AC2(a), since
$(M,\vec{u}) \sat [X_1 \gets x_1', \vec{X}_{-1} \gets \vec{x}'_{-1},
  \vec{W} \gets \vec{w}]\neg \phi$ by assumption.
AC3 trivially holds for $X_1=x_1$, so it remains to deal with
AC2(b).
Suppose, by way of contradiction, that $(M,\vec{u}) \sat [X_1 \gets
  x_1, \vec{X}_{-1} 
  \gets \vec{x}_{-1}', \vec{W} \gets \vec{w}]\neg \phi$.  
This means that $\vec{X}_{-1} \gets \vec{x}_{-1}$ satisfies
AC2(a$^m$), showing that AC3 (more precisely, the version of AC3
appropriate for the modified HP definition) is violated
(taking $(\<X_1\> \cdot \vec{W}, \<x_1\>
\cdot \vec{w}, \vec{x}_{-1}')$ as the witness),
 and $\vec{X}
\gets \vec{x}$ is \emph{not} a cause of $\phi$ in $(M,\vec{u})$
according to the modified HP definition, a
contradiction.  Thus, $(M,\vec{u}) \sat [X_1 \gets x_1, \vec{X}_{-1}
  \gets \vec{x}_{-1}', \vec{W} \gets \vec{w}] \phi$.
%Since all the variables in $\vec{W}$ are already
%set to their original values in $\vec{w}$ (i.e., their values in
%$(M,\vec{u})$), and all the orignal values of the variables in
%$\vec{X}^*$ are given by by $\vec{x}$, and hence, by AC1 and the
%fact that $\vec{X} = \vec{x}$ is a cause of $\phi$ according to the
%original values 

This does not yet show that AC2(b) holds: there might be
some subset $\vec{Z}'$ of variables in $\V - \vec{X}_{-1} \union \vec{W}$
that change value when $\vec{W}$ is set to $\vec{w}$ and
$\vec{X}_{-1}$ is set to $\vec{x}_{-1}$, and when these variables are
set to their original value in $(M,\vec{u})$, $\phi$ does not
hold, thus violating AC2(b).  More precisely, suppose that there
exists $\vec{Z}' = \<Z_1, \ldots, Z_k\> \subseteq \vec{Z}$ and 
values $z_j \ne  
z_j'$ for each variable $Z_j \in \vec{Z}'$ such that 
(i) $(M,\vec{u}) \sat Z_j = z_j$, 
(ii) $(M,\vec{u}) \sat [X_1 \gets x_1, \vec{X}_{-1} \gets
  \vec{x}'_{-1}, \vec{W} \gets \vec{w}](Z_j = z_j')$, and
(iii) $(M,\vec{u}) \sat [X_1 \gets x_1, \vec{X}_{-1} \gets
  \vec{x}'_{-1}, \vec{W} \gets \vec{w}, \vec{Z}' = \vec{z}]\neg \phi$.
But then $\vec{X} = \vec{x}$ is not a
cause of $\phi$ in $(M,\vec{u})$ according the modified HP definition.
 Condition
(iii) shows that AC2(a$^m$) is satisfied for $\vec{X}_{-1}$,
taking $(\<X_1\> \cdot \vec{W} \cdot \vec{Z}', \<x_1\>
\cdot \vec{w} \cdot \vec{z}, \vec{x}_{-1}')$ as the witness,
so again, AC3 is violated. It follows that AC2(b) holds.
Thus, $X=x$ is a cause of $\phi$ in $(M,\vec{u})$ according to
the original HP definition.   

The argument that $X=x$ is a cause of $\phi$ in $(M,\vec{u})$ according to
the updated HP definition is similar in spirit.  Indeed,
we just need to show one more thing.  For AC2(b$^u$), we must show
that if $\vec{X}' \subseteq \vec{X}_{-1}$, $\vec{W}' \subseteq
\vec{W}$, and $\vec{Z}' \subseteq \vec{Z}' \subseteq \vec{Z}$,
%$\vec{X}'' = \vec{X}_{-1} - \vec{X}'$,  
then 
\begin{equation}\label{eq:connected}
(M,\vec{u}) \sat [X_1 \gets x_1, \vec{X}' \gets \vec{x}',
\vec{W}' \gets \vec{w}, \vec{Z}' = \vec{z}] \neg \phi.
\end{equation}
(Here I am using the abuse of notation that I referred to in
Section~\ref{sec:causalitydef}, where if $\vec{X}' \subseteq \vec{X}$ and
$\vec{x} \in \R(\vec{X})$, I write $\vec{X}' \gets \vec{x}$, with the
intention that the components of $\vec{x}$ not included in $\vec{X}'$
are ignored.)
It follows easily from AC1 that 
(\ref{eq:connected}) holds if $\vec{X}' = \emptyset$.
And if (\ref{eq:connected}) does not hold for some strict nonempty subset
$\vec{X}'$ of $\vec{X}_{-1}$, then $\vec{X} =
\vec{x}$ is not a cause of $\phi$ according to the modified HP
definition because AC3 does not hold; AC2(a$^m$) is satisfied for
$\vec{X}'$. 
\eprf
}

\paragraph{Acknowledgments} I thank Sander Beckers, Hana Chockler,
Chris Hitchcock, and Joost Vennekens for useful comments and discussions.

\bibliographystyle{named}
\bibliography{z,joe}
\end{document}